\documentclass[journal]{IEEEtai}

\usepackage[colorlinks,urlcolor=blue,linkcolor=blue,citecolor=blue]{hyperref}

\usepackage{color,array}
\usepackage{graphicx}
\usepackage{amsmath}
\usepackage{xcolor}
\usepackage{longtable}
\usepackage{caption}
\usepackage{hyperref}
\usepackage{academicons}
\usepackage{caption}
\usepackage{multirow}
\usepackage{adjustbox}
\usepackage{rotating}
\usepackage{amsmath}
\usepackage{pgfplots}
\pgfplotsset{compat=1.18,width=0cm}
\usepackage{neuralnetwork}
\usepackage{tikz}
\usepackage{CJKutf8}
\usepackage{tcolorbox}
\usepackage[square,sort,comma,numbers]{natbib}
\usepackage{amssymb}
\usepackage{tabularx}
\usepackage{graphicx,subcaption}
\usepackage{lipsum}
\usepackage{authblk}

\usepackage{algorithm}
\usepackage{algpseudocode}

\begin{document}


\title{Enhancing Time Series Forecasting with Fuzzy Attention-Integrated Transformers}

\author{Sanjay Chakraborty, Fredrik Heintz}

\affil[]{Department of Computer and Information Science (IDA), REAL, AIICS, Linköping University, Linköping, Sweden, Email: sanjay.chakraborty@liu.se} 

\maketitle

\begin{abstract}
This paper introduces FANTF (Fuzzy Attention Network-Based Transformers), a novel approach that integrates fuzzy logic with existing transformer architectures to advance time series forecasting, classification, and anomaly detection tasks. FANTF leverages a proposed fuzzy attention mechanism incorporating fuzzy membership functions to handle uncertainty and imprecision in noisy and ambiguous time series data. The FANTF approach enhances its ability to capture complex temporal dependencies and multivariate relationships by embedding fuzzy logic principles into the self-attention module of the existing transformer's architecture. The framework combines fuzzy-enhanced attention with a set of benchmark existing transformer-based architectures to provide efficient predictions, classification and anomaly detection. Specifically, FANTF generates learnable fuzziness attention scores that highlight the relative importance of temporal features and data points, offering insights into its decision-making process. Experimental evaluatios on some real-world datasets reveal that FANTF significantly enhances the performance of forecasting, classification, and anomaly detection tasks over traditional transformer-based models. 
\end{abstract}

\textbf{Keywords -} Time-series; Forecasting; Anomaly detection; Classification, Transformer; Fuzzy attention network.

\section{Introduction}
Time-series forecasting performs a vital role across numerous industries, including sensor network tracking \citep{li2024deep}, automatic grid management \citep{maurya2016time}, finance industry \citep{zhang2024hybrid}, medical data analysis \citep{wen2022transformers}, anomaly detection in industrial maintenance \citep{xu2021anomaly}, and trajectory classification for action recognition \citep{franceschi2019unsupervised}. Traditionally dominated by Long Short-Term Memory (LSTM) and Recurrent Neural Networks (RNN), time-series data processing has witnessed a paradigm shift with the advent of transformers leveraging self-attention mechanisms \citep{atabay2022multivariate}. Initially revolutionizing natural language processing, transformers have now demonstrated superior performance in time-series analysis. These models excel in isolating and analyzing frequency components to detect trends and seasonality while effectively capturing both sequential dynamics and multivariate relationships \citep{atabay2022multivariate, ahmed2023transformers}. The integration of attention mechanisms enables transformers to model pairwise temporal dependencies among data points, making them highly effective for capturing complex patterns in sequential data \citep{li2019enhancing, kitaev2020reformer, zhou2021informer}. Consequently, transformer-based architectures have gained significant traction in time-series forecasting and sequential modeling \citep{dosovitskiy2020image, liu2021swin}, marking a transformative step in the field.
\par This paper aims to advance beyond traditional full attention mechanisms by introducing and exploring the learnable fuzzy self-attention (FAN) mechanism for time-series prediction. The primary objective is to enable the model to automatically discover an optimal balance between computational efficiency and predictive performance. By incorporating learnable fuzziness into the self-attention framework, the proposed approach seeks to enhance the model's ability to capture complex temporal dependencies, making it particularly effective for forecasting, classification and anomaly detection of time series data. The main contributions of this paper are as follows:
\begin{itemize}
    \item We have introduced a fuzzy self-attention network (FAN) for the existing benchmark transformer models and described its working procedure for multivariate time series analysis.
    \item We have described how FAN integrates into the full attention layer of existing transformer architectures, enabling efficient multivariate time series forecasting, classification, and anomaly detection tasks.
    \item FANTF is a lightweight and computationally efficient approach that integrates with existing transformer models, improving accuracy in forecasting, classification, and anomaly detection tasks while outperforming state-of-the-art models.
    \item We conduct comprehensive evaluations across diverse time-series datasets. Experimentally, FANTF obtains notable performance improvement compared to some existing benchmark models, underscoring its effectiveness. Our in-depth analysis of its embedded modules and architectural decisions highlights promising directions for advancing transformer-based forecasting models, paving the way for further innovations in this domain. 
\end{itemize}

\section{Background}
\label{background}
\subsection{Transformers for Time Series}
In this section, we briefly discuss prominent transformer models that are highly effective for tasks like time-series forecasting, classification, and anomaly detection. Transformer models have gained substantial attention in both short-term and long-term time-series forecasting due to their ability to capture intricate temporal dependencies \cite{zerveas2021transformer, zhang2024multivariate, nie2022time, zeng2023transformers}. Among early advancements, Informer \citep{zhou2021informer} stands out for introducing a generative-based decoder and 'Probability-Sparse' self-attention, which deal with the problem of quadratic time complexity. Building on this, models such as Autoformer \citep{chen2021autoformer}, iTransformer \citep{liu2023itransformer}, and FEDFormer \citep{zhou2022fedformer} have emerged. iTransformer \citep{liu2023itransformer} innovates by embedding individual time points as variate tokens, leading the attention mechanism to model multivariate correlations while using feed-forward networks to learn nonlinear representations. PatchTST \citep{nie2022time} improves local and global dependency capture through patch-based processing. Crossformer \citep{zhang2023crossformer} introduces a dimension-segment-wise (DSW) integrating technique that encodes time-series data into a 2-dimensional array of vectors. The core of transformer models is their powerful attention mechanism, which enables them to focus on relevant parts of the input sequence for accurate predictions \citep{zeng2023transformers}. By computing weighted representations through attention scores between query, key, and value vectors, transformers effectively capture dependencies across sequence positions, irrespective of their distance. Scaled dot-product attention ensures stable gradients, while multi-head attention extends this by learning diverse patterns from different input subspaces. These capabilities make transformers particularly suited for sequential and structured data tasks, including natural language processing and time-series analysis.

\subsection{Fuzzy Logic and Fuzziness}
Fuzzy logic and membership functions can enhance transformer architectures by incorporating the ability to handle uncertainty and imprecision in data. It can be extensively used with deep learning architectures for time series analysis \citep{zhan2023differential}. In transformers, fuzzy membership functions can be applied to define degrees of relevance or importance for input features, rather than relying on binary or crisp classifications. This approach enables the attention mechanism to weigh relationships more flexibly, reflecting the nuanced and uncertain nature of real-world data. By embedding fuzzy logic into transformers, the models can better manage noise, ambiguous patterns, and incomplete information, making them more robust and interpretable, particularly in domains like time series analysis, forecasting, and anomaly detection \citep{engel2024transformer}. The membership function \( \mu_A(x) \) defines the degree of membership of \( x \) in a fuzzy set \( A \), where \( \mu_A(x) \in [0,1] \):
\begin{equation}
\mu_A(x) : X \to [0,1]
\end{equation}
We have briefly discussed here some interesting membership functions. \\
1. \textbf{Gaussian Fuzziness}: Gaussian fuzziness refers to the uncertainty modeled by a Gaussian function, often used to represent membership grades in fuzzy sets. The degree of membership \( \mu_G(x) \) follows a bell-shaped curve, where values near the mean have higher membership, and the membership decreases as \( x \) moves further from the mean. This approach captures smooth transitions between membership states.
\begin{equation}
\mu_G(x) = \exp\left( -\frac{(x - c)^2}{2 \sigma^2} \right)
\end{equation}
Where, c is the center (mean) of the Gaussian function and $\sigma$ is the standard deviation that controls the width of the Gaussian curve.

2. \textbf{Scaled Sigmoid}: A scaled sigmoid function is a variant of the standard sigmoid function, \( \sigma(x) = \frac{1}{1 + e^{-x}} \), where the output is adjusted by scaling factors to fit specific ranges. This function is often used in fuzzy systems to model a smooth, continuous transition between two states, with its slope and range being adaptable to different applications.

3. \textbf{Learnable Fuzziness}: Learnable fuzziness refers to the concept where the degree of fuzziness in a system is not fixed but is instead learned through data-driven methods. By training a model, parameters that define the fuzziness can be adjusted to optimize system performance, allowing for adaptive behavior based on observed patterns.
\begin{equation}
\mu_A(x; \theta) = \sigma\left( \frac{x - \theta_1}{\theta_2} \right)
\end{equation}
\(\mu_A(x; \theta)\) represents the membership degree of \( x \) in the fuzzy set \( A \), where \( \theta_1 \) and \( \theta_2 \) are learnable parameters (e.g., the center and scale of a sigmoid function). The sigmoid function is defined as:
\[
\sigma(z) = \frac{1}{1 + e^{-z}}
\]
The parameters \( \theta_1 \) and \( \theta_2 \) are adjusted during the training process, allowing the fuzziness (i.e., the transition between membership values) to adapt based on the data.

4. \textbf{Uniform Distribution}: A uniform distribution is a probability distribution where every outcome in a given range has the same likelihood of occurring. In fuzzy systems, it can be used to model uncertainty when all values within a range are equally probable, providing a simple yet effective representation of randomness or lack of prior knowledge.
\begin{equation}
f_X(x) = 
\begin{cases} 
\frac{1}{b - a} & \text{if } a \leq x \leq b, \\
0 & \text{otherwise},
\end{cases}
\end{equation}

In this work, we employed a learnable fuzziness technique to compute the fuzziness score for the attention mechanism. Learnable fuzziness offers significant advantages over fixed methods like Gaussian, scaled sigmoid, or uniform fuzziness by dynamically adapting to data and task-specific requirements during training. Unlike predefined approaches, which apply a fixed degree of uncertainty, learnable fuzziness optimizes the level of noise or variation incorporated, enabling the model to balance robustness and sensitivity to patterns in diverse data scenarios. This adaptability allows it to handle noisy, irregular, or sparse datasets more effectively, capturing complex, non-linear relationships that static fuzziness methods may miss. Additionally, learnable fuzziness enhances interpretability by providing insights into how the model adjusts uncertainty to improve attention weights, resulting in better generalization and performance across tasks like forecasting, classification, and anomaly detection.

\section{Problem Statements}
\label{prob}
This work deals with the challenges of long-term and short-term multivariate time series (MTS) forecasting by utilizing historical data while also considering classification and anomaly detection tasks. A MTS at time t is defined as $({X}_t = [x_{t,1}, x_{t,2}, \dots, x_{t,N}])$, where $x_{t,n}$ denotes the value of the n-th variable at time t for n = 1, 2,...., N. The notation ${X}_{t:t+H}$ is used to represent the series values from time t to t+H, inclusive. However, for a given starting time $t_0$, the model receives as input the sequence ${X}_{t_0-L:t_0}$, representing the last L time steps, and produces the predicted sequence \( \hat{\mathbf{X}}_{t_0:t_0+H} \), corresponding to the forecasted values for the following \( H \) time steps. The forecasted value at any time t is denoted as \( \hat{\mathbf{X}}_t \). 

\begin{equation}
\hat{\mathbf{X}}_{t:t+H} = f(\mathbf{X}_{t-L:t})
\end{equation}

Given a time series dataset \( X = \{ X_1, X_2, \dots, X_N \} \), where \( X_i = [ x_{i,1}, x_{i,2}, \dots, x_{i,T} ] \) represents a sequence of observations, the objective is to assign each sequence \( X_i \) to one of \( C \) possible classes \( \{ y_1, y_2, \dots, y_C \} \). The challenge lies in capturing both global and local temporal dependencies, handling noisy and irregular data, and ensuring robustness across various time series lengths. Applications span diverse domains, including healthcare, finance, and activity recognition, where accurate classification is critical for decision-making.
\begin{equation}
y_i = \arg\max_{c \in \{y_1, y_2, \dots, y_C\}} P(y = c \mid X_i)
\end{equation}
where \( P(y = c \mid X_i) \) is the probability of class \( c \) given the input time series \( X_i \), and the objective is to assign the label \( y_i \) to the class with the highest probability.

Time series anomaly detection aims to find irregular patterns within temporal data that deviate from normal behavior. Given a time series \( X = [ x_1, x_2, \dots, x_T ] \), the goal is to detect instances \( t \) where \( x_t \) or a segment \( X_{t:t+k} \) exhibits anomalies. These anomalies may arise due to faults, unusual events, or rare occurrences, and their detection is crucial in applications such as system monitoring, fraud detection, and predictive maintenance. The task is complicated by the need to distinguish genuine anomalies from noise, adapt to non-stationary data, and minimize false positives while ensuring timely detection.
\begin{equation}
\mathcal{A} = \{ t : |x_t - \hat{x}_t| > \epsilon \}
\end{equation}
where \( \hat{x}_t \) is the predicted value of \( x_t \) based on past observations, and \( \epsilon \) is a predefined threshold that determines if the deviation is considered an anomaly.

\section{Methodology}
\label{Method}
\begin{figure*}[hbt!]
\centering
\includegraphics[scale=0.35]{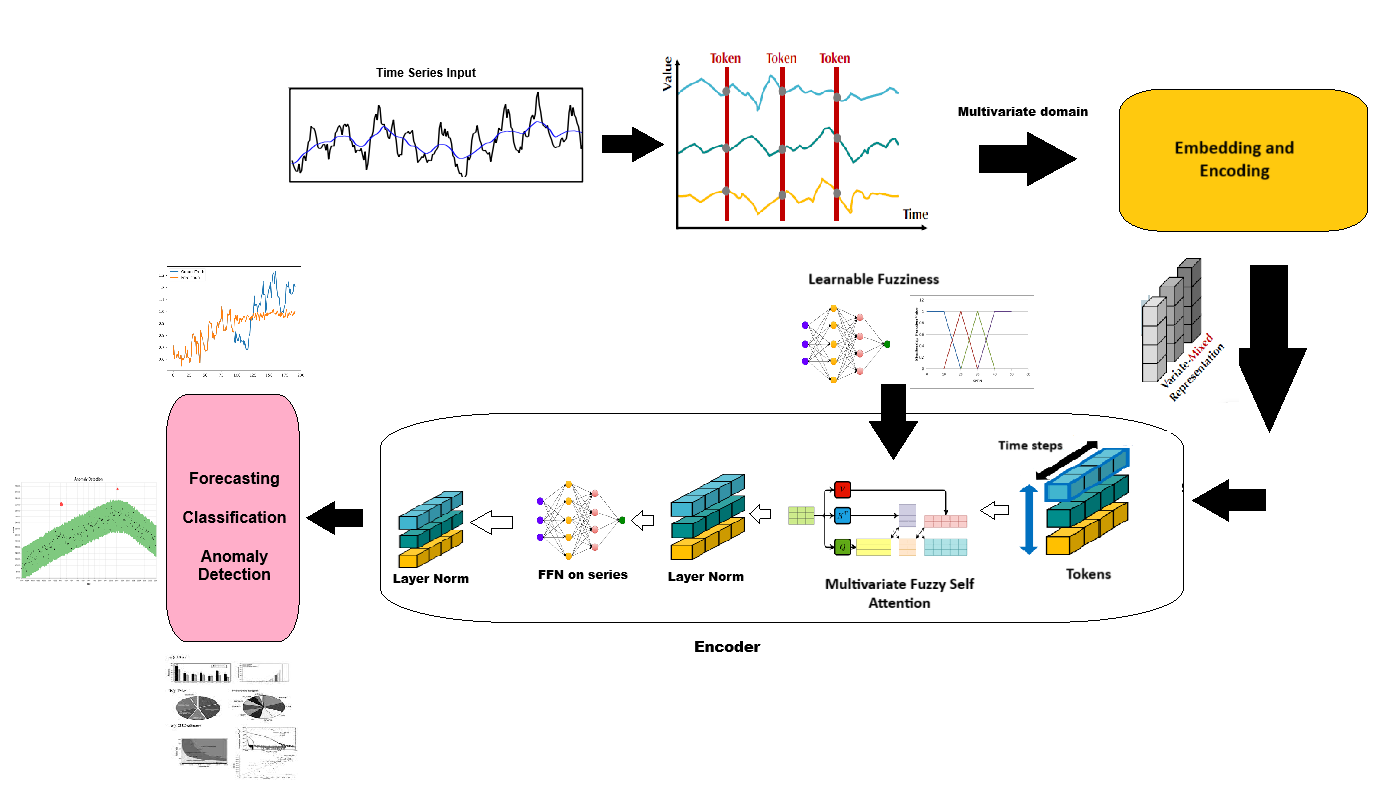}
\caption{Overall approach of FANTF}
\label{FANTF_architecture}
\end{figure*}

In this work, we have explained the proposed FAN mechanism integrated in some existing state-of-the-art time series transformer architectures, which include an input embedding operation, fuzzy self-attention network (FAN) design, layer normalization, and feed-forward networks (FFNs). The general workflow of the FANTF approach is depicted in Figure \ref{FANTF_architecture}. Incorporating learnable fuzziness into fuzzy attention networks with transformers enhances their adaptability and performance compared to traditional attention mechanisms. The learnable fuzziness parameter allows the model to dynamically adjust the level of uncertainty it incorporates, enabling it to better handle diverse and noisy data scenarios. This flexibility improves the network's ability to capture nuanced relationships and subtle temporal or spatial patterns that might be overlooked by standard attention mechanisms. Additionally, the learnability of fuzziness fosters improved interpretability by providing insights into the model's confidence and focus, while maintaining robustness across tasks like forecasting, classification, and anomaly detection. This approach leverages the strengths of transformers while mitigating their susceptibility to overfitting or over-reliance on specific patterns, offering a more generalized and effective solution for complex, real-world datasets.

\subsection{Model Inputs}
\begin{figure}[hbt!]
\centering
\includegraphics[scale=0.35]{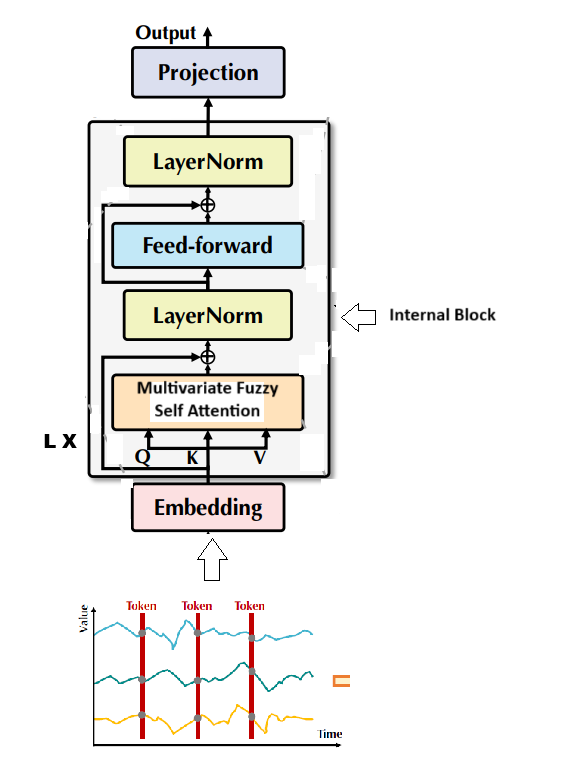}
\caption{Architecture of FANTF internal blocks}
\label{Fig2}
\end{figure}

For time series applications, the input to a Transformer model is structured as a sequence of observations, $ X = [x_1, x_2, \dots, x_T] $, where T is the sequence length. Each observation can be a scalar or a multivariate vector, depending on the number of features. The data is often normalized and embedded into a higher-dimensional space using linear layers or embeddings. To preserve temporal order, positional encodings are added to the embeddings, providing the model with information about the sequence's structure. The resulting input tensor, typically of shape (B, T, D), where B is the batch size and D is the feature dimension, is fed into the transformer's encoder for processing. This design enables the model to capture both local and global dependencies within the time series.
\begin{equation}
\begin{split}
h^0_n=Embedding(X:,n)\\
H^{(l+1)}=IntBlock(H^l), l= 0, ....., L-1, \\
Y^t:,n = Projection(h^L_n),
\end{split}
\end{equation}
where the superscript denotes the layer index and \( H = \{h_1, \dots, h_N \} \in \mathbb{R}^{N \times D} \) contains N embedded tokens of size D. Multi-layer perceptrons (MLPs) are employed to implement both the embedding and projection. The shared feed-forward network within each \texttt{IntBlock()} processes the acquired tokens individually while enabling interaction between them via fuzzy self-attention. The internal block architecture of FANTF is shown in Figure \ref{Fig2}.

\subsection{Encoding of Model}
In this block, we have organized a stack of L number of blocks, each consisting of the proposed fuzzy attention network (FAN), feed-forward network, and layer normalization modules.

\subsubsection{Proposed Fuzzy Attention (FAN)}
The fuzzy attention mechanism is commonly employed to model temporal connections in tasks such as forecasting, classification, and anomaly detection. The self-attention mechanism enables the model to assign weights to different tokens in a sequence, facilitating the capture of both long-range and short-range dependencies and contextual relationships. This attention module is designed to be applicable to both the encoder and decoder architectures of a transformer. It utilizes linear projections to receive queries (\(Q\)), keys (\(K\)), and values (\(V\)):
\begin{equation}
Q = SW_Q, K = SW_k, V = SW_v
\end{equation}
where $W_Q$, $W_k$, and $W_v$ are learnable weight matrices. Multi-head attention concatenates the attention results to generate the desired output after multiple keys, queries, and values are entered into multi-scaled dot-product attention blocks. The attention scores are evaluated as:
\begin{equation}
Scores = \frac{QK^T}{\sqrt{d_k}} + \delta . \mathcal{N}(0, \sigma^2)
\end{equation}
where n defines the sequence length of input, \(d_k\) denotes the projected dimension, \( \delta \) is the learnable fuzziness parameter, and \( \mathcal{N}(0, \sigma^2) \) represents Gaussian noise. As a result, we perform Softmax normalization $(O(n^2 d))$ and weighted sum operations.\\\\
Attention(Q, K, V) = Softmax(Scores)\\\\
Output = Attention(Q, K, V) . V \\\\
Rather than using a single attention function, the Transformer employs multivariate self-attention (MVA) with multiple sets of learned projections (H distinct time series).
\begin{equation}
MVA(Q, K, V) = Concate(h_1,......., h_H) W^O
\end{equation}
where $(h_i = Attention(Q W_i^Q, K W_i^K, V W_i^V)$. The learnable fuzziness in attention is described in Algorithm 1.

\subsubsection{Normalization of Layers}
The next block is "Layer Normalization (LayerNorm)," which was originally proposed to improve the stability and convergence of deep networks during training. This module, which gradually integrates variables, normalizes the multivariate representation of the same timestamp in the majority of transformer-based forecasters. Nevertheless, as explained in Equation \ref{eq11}, our inverted design normalizes the series representation of individual variates. It has been demonstrated that this strategy works well for non-stationary situations. On the contrary, an oversmoothed time series is frequently the result of the prior approach, which normalizes across several time-step tokens.
\begin{equation}
\label{eq11}
LayerNorm(H) = \left[  \frac{ h_n - Mean(h_n) } { \sqrt{Var(h_n)} } | n = 1, ....., N \right]
\end{equation}

\subsubsection{Feed-forward network}
The Transformer's fundamental component for token representation encoding is the feed-forward network (FFN), which is applied uniformly to each token. Multiple timestamp variants that make up the token in the vanilla Transformer could be mismatched and excessively localized, which leaves out important information for precise forecasts. On the other hand, each variate token's series representation is subjected to the FFN. With this method, the observed decomposed time series is stored, and blocks are stacked to decode the representations for following series using dense non-linear connections. 
\begin{equation}
FFN(H')=ReLU(H'W^1+b^1)W^2+b^2
\end{equation}
Where, \( H' \) is the output of the previous layer, and \( W_1, W_2, b^2 \) are trainable parameters.

\vfill
\begin{algorithm}[H]
\caption{Learnable Fuzziness in Multi-Head Attention}
\begin{algorithmic}
\State \textbf{Input:} Queries \( Q \), Keys \( K \), Values \( V \), Fuzziness Parameter \( \delta \), Learned Weight Matrices \( W_Q, W_K, W_V \), Sequence Length \( n \), Projected Dimension \( d_k \), Gaussian Noise \( \mathcal{N}(0, \sigma^2) \)
\State \textbf{Output:} Attention Output \( \text{Output} \)
\vspace{2mm}

\Procedure{LearnableFuzzinessInMHA}{$Q, K, V, \delta, W_Q, W_K, W_V, n, d_k, \mathcal{N}(0, \sigma^2)$}
    \State \textbf{Initialize:} Weight matrices \( W_Q, W_K, W_V \) as learned parameters
    \State \textbf{Linear projections:} 
    \[
    Q = S W_Q, \quad K = S W_K, \quad V = S W_V
    \]
    \State Compute the dot product attention scores: 
    \[
    \text{scores} = \frac{Q K^T}{\sqrt{d_k}} + \delta \cdot \mathcal{N}(0, \sigma^2)
    \]
    \If{attn\_mask is not provided}
        \State Apply causal mask to the scores
    \EndIf
    \State Apply Softmax normalization: 
    \[
    A = \text{Softmax}(\text{scores})
    \]
    \State Apply dropout to attention weights \( A \)
    \State Compute the output using the attention weights: 
    \[
    \text{Output} = A \cdot V
    \]
\vspace{2mm}
    \State \textbf{Multi-Head Attention (MHA):}
    \For{$i = 1$ to \( H \)} 
        \State Compute each head \( h_i = \text{Attention}(Q W_i^Q, K W_i^K, V W_i^V) \)
    \EndFor
    \State Concatenate the attention heads: 
    \[
    \text{MVA}(Q, K, V) = \text{Concate}(h_1, \dots, h_H) W^O
    \]
    \State \textbf{Return:} Output from multi-head attention: \( \text{MVA}(Q, K, V) \)
\EndProcedure
\end{algorithmic}
\end{algorithm}

\begin{equation}
H'=LayerNorm(MVSelfAtten(X)+X)
\end{equation}
\begin{equation}
H=LayerNorm(FFN(H')+H')
\end{equation}
Where, \( \text{MVSelfAtten}(\cdot) \) denotes the self-attention module for multivariate and \( \text{LayerNorm}(\cdot) \) defines the layer normalization job.

\section{Result Analysis}
\label{result}
\subsection{Datasets}
The datasets comprise both long- and short-term time series forecasting, classification, and anomaly detection scenarios in this paper. All these datasets are described in detail in the supplementary section. In this section, we compare FANTF's performance on all three tasks to that of typical transformers.

\subsection{Long- and Short-term Forecasting}
We have provided comprehensive experiments in this area to assess the effectiveness of our suggested FANTF approach in comparison to the most advanced time-series forecasting methods. To evaluate the effect of the proposed modules, we have also carried out an ablation investigation. Every experiment is carried out on a single NVIDIA-GeForce RTX 3090 GPU using PyTorch and CUDA version 12.2. Using the Time-Series Library (TSLib) repository, we have reproduced baseline models that match setups found in the official code and documentation. The introduction of a fuzzy attention module significantly enhances the model's performance. Our FANTF approach consistently outperforms state-of-the-art models such as Informer, Crossformer, iTransformer, vanilla Transformer, and PatchTST. Figure \ref{predtest1} shows a sample long-term prediction comparison for these models and their FAN versions on the Exchange dataset. The performance comparison is further illustrated in Figure \ref{fig-comparison}, with individual prediction horizon results provided in supplementary tables.
Table \ref{avgtab} compares multivariate long-term forecasting results across various datasets, using models like iTransformer, Informer, Transformer, PatchTST, and Crossformer, both with and without FAN. The results, based on MSE and MAE, highlight the impact of FAN, with improvements shown in blue. These results demonstrate FAN's ability to enhance forecasting accuracy across most datasets. For short-term forecasting, the FANTF approach shows competitive performance improvements compared to iTransformer, Informer, Crossformer, and PatchTST, as shown in Table \ref{avgtabshort}. Additional results from the M4 dataset in Table \ref{table-resultsM4} further validate the effectiveness of FANTF. The sample short-term predictions on the PEMS08 dataset, shown in Figure \ref{fig-PEMS08-compare}, highlight the competitive accuracy of FANTF compared to state-of-the-art models. Furthermore, FAN's lightweight design ensures it delivers comparable results in less time, making it an efficient augmentation to existing models.

\begin{table*}[hbt!]
\centering
\scriptsize
\captionsetup{justification=centering}
\caption{An overview of the benchmark datasets' outcomes. Before and after using FAN, average error coefficients (MSE, MAE) on multivariate long-term forecasting results are compared. The difference between the results is displayed in \%, with the blue colors signifying that FANTF is superior to the current models. The negative growths are represented by hyphen.}
\begin{tabular}{|c|cc|cc|cc|cc|cc|cc|cc|cc|}
\hline
Database                 & \multicolumn{2}{c|}{ETTh1}                                                                       & \multicolumn{2}{c|}{ETTh2}                                                                         & \multicolumn{2}{c|}{ETTm1}                                                                       & \multicolumn{2}{c|}{ETTm2}                                                                        & \multicolumn{2}{c|}{Weather}                                                                      & \multicolumn{2}{c|}{Electricity}                                                                  & \multicolumn{2}{c|}{Traffic}                                                                       & \multicolumn{2}{c|}{Exchange}                                                                      \\ \hline
Models                   & MSE                                                       & MAE                                  & MSE                                                        & MAE                                   & MSE                                                       & MAE                                  & MSE                                                        & MAE                                  & MSE                                                       & MAE                                   & MSE                                                       & MAE                                   & MSE                                                        & MAE                                   & MSE                                                        & MAE                                   \\ \hline
iTransformer             & 0.457                                                     & 0.448                                & 0.396                                                      & 0.413                                 & 0.410                                                     & 0.411                                & 0.294                                                      & 0.336                                & 0.252                                                     & 0.282                                 & 0.188                                                     & 0.277                                 & 0.443                                                      & 0.299                                 & 0.377                                                      & 0.418                                 \\ \hline
iTransformer+FAN         & 0.456                                                     & 0.446                                & 0.387                                                      & 0.408                                 & 0.409                                                     & 0.409                                & 0.291                                                      & 0.334                                & 0.251                                                     & 0.279                                 & 0.183                                                     & 0.272                                 & 0.440                                                      & 0.294                                 & 0.361                                                      & 0.404                                 \\ \hline
\textbf{Difference (\%)} & \multicolumn{1}{c|}{{\color[HTML]{3166FF} \textbf{0.22}}} & {\color[HTML]{3166FF} \textbf{0.45}} & \multicolumn{1}{c|}{{\color[HTML]{3166FF} \textbf{2.27}}}  & {\color[HTML]{3166FF} \textbf{1.21}}  & \multicolumn{1}{c|}{{\color[HTML]{3166FF} \textbf{0.24}}} & {\color[HTML]{3166FF} \textbf{0.49}} & \multicolumn{1}{c|}{{\color[HTML]{3166FF} \textbf{1.02}}}  & {\color[HTML]{3166FF} \textbf{0.60}} & \multicolumn{1}{c|}{{\color[HTML]{3166FF} \textbf{0.40}}} & {\color[HTML]{3166FF} \textbf{1.06}}  & \multicolumn{1}{c|}{{\color[HTML]{3166FF} \textbf{2.66}}} & {\color[HTML]{3166FF} \textbf{1.81}}  & \multicolumn{1}{c|}{{\color[HTML]{3166FF} \textbf{1.81}}}  & {\color[HTML]{3166FF} \textbf{1.67}}  & \multicolumn{1}{c|}{{\color[HTML]{3166FF} \textbf{0.68}}}  & {\color[HTML]{3166FF} \textbf{3.35}}  \\ \hline
Informer                 & 1.058                                                     & 0.808                                & 4.665                                                      & 1.772                                 & 0.890                                                     & 0.702                                & 1.716                                                      & 0.903                                & 0.628                                                     & 0.548                                 & 0.362                                                     & 0.439                                 & 0.862                                                      & 0.487                                 & 1.622                                                      & 1.005                                 \\ \hline
Informer+FAN             & 0.964                                                     & 0.780                                & 3.249                                                      & 1.440                                 & 0.854                                                     & 0.702                                & 1.553                                                      & 0.873                                & 0.633                                                     & 0.551                                 & 0.278                                                     & 0.373                                 & 0.659                                                      & 0.355                                 & 1.438                                                      & 0.931                                 \\ \hline
\textbf{Difference (\%)} & \multicolumn{1}{c|}{{\color[HTML]{3166FF} \textbf{8.88}}} & {\color[HTML]{3166FF} \textbf{3.47}} & \multicolumn{1}{c|}{{\color[HTML]{3166FF} \textbf{30.3}}}  & {\color[HTML]{3166FF} \textbf{18.74}} & \multicolumn{1}{c|}{{\color[HTML]{3166FF} \textbf{4.04}}} & {\color[HTML]{000000} \textbf{0}}    & \multicolumn{1}{c|}{{\color[HTML]{3166FF} \textbf{9.50}}}  & {\color[HTML]{3166FF} \textbf{3.32}} & \multicolumn{1}{c|}{{\color[HTML]{3166FF} \textbf{-}}}    & {\color[HTML]{3166FF} \textbf{-}}     & \multicolumn{1}{c|}{{\color[HTML]{3166FF} \textbf{-}}}    & {\color[HTML]{3166FF} \textbf{36.67}} & \multicolumn{1}{c|}{{\color[HTML]{3166FF} \textbf{36.67}}} & {\color[HTML]{3166FF} \textbf{27.10}} & \multicolumn{1}{c|}{{\color[HTML]{3166FF} \textbf{59.37}}} & {\color[HTML]{3166FF} \textbf{7.36}}  \\ \hline
Transformer              & 0.952                                                     & 0.777                                & 4.528                                                      & 1.689                                 & 0.961                                                     & 0.743                                & 1.511                                                      & 0.869                                & 0.631                                                     & 0.568                                 & 0.267                                                     & 0.364                                 & 0.670                                                      & 0.367                                 & 1.339                                                      & 0.807                                 \\ \hline
Transformer+FAN          & 0.957                                                     & 0.780                                & 4.490                                                      & 1.681                                 & 0.948                                                     & 0.739                                & 1.507                                                      & 0.875                                & 0.605                                                     & 0.557                                 & 0.265                                                     & 0.362                                 & 0.670                                                      & 0.364                                 & 1.349                                                      & 0.887                                 \\ \hline
\textbf{Difference (\%)} & \multicolumn{1}{c|}{{\color[HTML]{3166FF} \textbf{-}}}    & {\color[HTML]{3166FF} \textbf{-}}    & \multicolumn{1}{c|}{{\color[HTML]{3166FF} \textbf{0.84}}}  & {\color[HTML]{3166FF} \textbf{0.47}}  & \multicolumn{1}{c|}{{\color[HTML]{3166FF} \textbf{1.35}}} & {\color[HTML]{3166FF} \textbf{0.54}} & \multicolumn{1}{c|}{{\color[HTML]{3166FF} \textbf{0.26}}}  & {\color[HTML]{3166FF} \textbf{-}}    & \multicolumn{1}{c|}{{\color[HTML]{3166FF} \textbf{4.12}}} & {\color[HTML]{3166FF} \textbf{4.12}}  & \multicolumn{1}{c|}{{\color[HTML]{3166FF} \textbf{1.94}}} & {\color[HTML]{3166FF} \textbf{0.75}}  & \multicolumn{1}{c|}{{\color[HTML]{3166FF} \textbf{0.75}}}  & {\color[HTML]{3166FF} \textbf{0.55}}  & \multicolumn{1}{c|}{{\color[HTML]{3166FF} \textbf{-}}}     & {\color[HTML]{3166FF} \textbf{-}}     \\ \hline
PatchTST                 & 0.458                                                     & 0.454                                & 0.394                                                      & 0.417                                 & 0.392                                                     & 0.407                                & 0.294                                                      & 0.338                                & 0.257                                                     & 0.279                                 & 0.214                                                     & 0.312                                 & 0.533                                                      & 0.342                                 & 0.394                                                      & 0.418                                 \\ \hline
PatchTST+FAN             & 0.455                                                     & 0.451                                & 0.391                                                      & 0.415                                 & 0.390                                                     & 0.406                                & 0.290                                                      & 0.336                                & 0.256                                                     & 0.277                                 & 0.210                                                     & 0.301                                 & 0.540                                                      & 0.350                                 & 0.383                                                      & 0.413                                 \\ \hline
\textbf{Difference (\%)} & \multicolumn{1}{c|}{{\color[HTML]{3166FF} \textbf{6.66}}} & {\color[HTML]{3166FF} \textbf{0.66}} & \multicolumn{1}{c|}{{\color[HTML]{3166FF} \textbf{0.76}}}  & {\color[HTML]{3166FF} \textbf{0.48}}  & \multicolumn{1}{c|}{{\color[HTML]{3166FF} \textbf{0.51}}} & {\color[HTML]{3166FF} \textbf{0.25}} & \multicolumn{1}{c|}{{\color[HTML]{3166FF} \textbf{1.36}}}  & {\color[HTML]{3166FF} \textbf{0.59}} & \multicolumn{1}{c|}{{\color[HTML]{3166FF} \textbf{0.39}}} & {\color[HTML]{3166FF} \textbf{0.72}}  & \multicolumn{1}{c|}{{\color[HTML]{3166FF} \textbf{-}}}    & {\color[HTML]{000000} \textbf{0}}     & \multicolumn{1}{c|}{{\color[HTML]{3166FF} \textbf{3.53}}}  & {\color[HTML]{000000} \textbf{0}}     & \multicolumn{1}{c|}{{\color[HTML]{3166FF} \textbf{28.14}}} & {\color[HTML]{3166FF} \textbf{23.52}} \\ \hline
Crossformer              & 0.557                                                     & 0.537                                & 2.768                                                      & 1.323                                 & 0.592                                                     & 0.568                                & 1.377                                                      & 0.761                                & 0.265                                                     & 0.327                                 & 0.278                                                     & 0.340                                 & 0.563                                                      & 0.306                                 & 0.938                                                      & 0.739                                 \\ \hline
Crossformer+FAN          & 0.552                                                     & 0.531                                & 2.291                                                      & 1.189                                 & 0.661                                                     & 0.599                                & 1.209                                                      & 0.730                                & 0.260                                                     & 0.312                                 & 0.179                                                     & 0.279                                 & 0.559                                                      & 0.305                                 & 0.881                                                      & 0.717                                 \\ \hline
\textbf{Difference (\%)} & \multicolumn{1}{c|}{{\color[HTML]{3166FF} \textbf{0.90}}} & {\color[HTML]{3166FF} \textbf{1.12}} & \multicolumn{1}{c|}{{\color[HTML]{3166FF} \textbf{17.23}}} & {\color[HTML]{3166FF} \textbf{10.13}} & \multicolumn{1}{c|}{{\color[HTML]{3166FF} \textbf{-}}}    & {\color[HTML]{3166FF} \textbf{-}}    & \multicolumn{1}{c|}{{\color[HTML]{3166FF} \textbf{12.20}}} & {\color[HTML]{3166FF} \textbf{4.07}} & \multicolumn{1}{c|}{{\color[HTML]{3166FF} \textbf{1.89}}} & {\color[HTML]{3166FF} \textbf{20.49}} & \multicolumn{1}{c|}{{\color[HTML]{3166FF} \textbf{4.59}}} & {\color[HTML]{3166FF} \textbf{35.61}} & \multicolumn{1}{c|}{{\color[HTML]{3166FF} \textbf{35.61}}} & {\color[HTML]{3166FF} \textbf{17.94}} & \multicolumn{1}{c|}{{\color[HTML]{3166FF} \textbf{6.08}}}  & {\color[HTML]{3166FF} \textbf{2.98}}  \\ \hline
\end{tabular}
\label{avgtab}
\end{table*}

\begin{table*}[hbt!]
\centering
\scriptsize
\captionsetup{justification=centering}
\caption{Comparison of average error metrics (MSE, MAE) on the PEMS dataset for multivariate short-term forecasting, using a prediction length of 48 and a look-back window of 96. Values highlighted in red denote the improvements achieved by FANTF over existing results.}
\begin{tabular}{|l|cc|cc|cc|cc|cc|cc|}
\hline
Models   & \multicolumn{2}{c|}{iTransformer+FAN}                                            & \multicolumn{2}{c|}{iTransformer}  & \multicolumn{2}{c|}{Crossformer+FAN}                                             & \multicolumn{2}{c|}{Crossformer}   & \multicolumn{2}{c|}{PatchTST+FAN}                                                & \multicolumn{2}{c|}{PatchTST}      \\ \hline
Database & MSE                                               & MAE                          & MSE                        & MAE   & MSE                                               & MAE                          & MSE                        & MAE   & MSE                                               & MAE                          & MSE                        & MAE   \\ \hline
PEMS03   & \multicolumn{1}{c|}{0.249}                        & 0.350                        & \multicolumn{1}{c|}{0.238} & 0.343 & \multicolumn{1}{c|}{0.175}                        & 0.286                        & \multicolumn{1}{c|}{0.157} & 0.273 & \multicolumn{1}{c|}{{\color[HTML]{FE0000} 0.300}} & {\color[HTML]{FE0000} 0.382} & \multicolumn{1}{c|}{0.309} & 0.383 \\ \hline
PEMS04   & \multicolumn{1}{c|}{0.218}                        & 0.319                        & \multicolumn{1}{c|}{0.218} & 0.319 & \multicolumn{1}{c|}{0.189}                        & 0.302                        & \multicolumn{1}{c|}{0.149} & 0.270 & \multicolumn{1}{c|}{{\color[HTML]{FE0000} 0.312}} & {\color[HTML]{FE0000} 0.386} & \multicolumn{1}{c|}{0.313} & 0.387 \\ \hline
PEMS07   & \multicolumn{1}{c|}{{\color[HTML]{FE0000} 0.231}} & {\color[HTML]{FE0000} 0.334} & \multicolumn{1}{c|}{0.244} & 0.349 & \multicolumn{1}{c|}{{\color[HTML]{FE0000} 0.175}} & {\color[HTML]{FE0000} 0.272} & \multicolumn{1}{c|}{0.179} & 0.273 & \multicolumn{1}{c|}{{\color[HTML]{FE0000} 0.292}} & {\color[HTML]{FE0000} 0.376} & \multicolumn{1}{c|}{0.298} & 0.379 \\ \hline
PEMS08   & \multicolumn{1}{c|}{{\color[HTML]{FE0000} 0.230}} & {\color[HTML]{FE0000} 0.316} & \multicolumn{1}{c|}{0.237} & 0.323 & \multicolumn{1}{c|}{{\color[HTML]{FE0000} 0.220}} & {\color[HTML]{FE0000} 0.274} & \multicolumn{1}{c|}{0.223} & 0.276 & \multicolumn{1}{c|}{0.272}                        & {\color[HTML]{FE0000} 0.349} & \multicolumn{1}{c|}{0.268} & 0.350 \\ \hline
\end{tabular}
\label{avgtabshort}
\end{table*}

\begin{table*}[hbt!]
\centering
\scriptsize
\captionsetup{justification=centering}
\caption{Summary of short-term forecasting results in the M4 dataset with a single variate. Every prediction length may be found in [6, 48]. The red-highlighted numbers show an improvement in FANTF above the current outcomes. The difference between the results is shown in \% where the blue colours represent improvements of FANTF over existing models. The negative growths are represented by a hyphen.}
\begin{tabular}{|l|c|cc|cc|cc|cc|}
\hline
\textbf{Metric}                  & \textbf{Category}                             & \multicolumn{1}{c|}{\textbf{iTransformer+FAN}}              & \textbf{iTransformer} & \multicolumn{1}{c|}{\textbf{PatchTST+FAN}}                  & \textbf{PatchTST} & \multicolumn{1}{c|}{\textbf{Informer+FAN}}        & \textbf{Informer} & \multicolumn{1}{c|}{\textbf{Crossformer+FAN}}               & \textbf{Crossformer} \\ \hline
                                 & Yearly                                        & \multicolumn{1}{c|}{{\color[HTML]{FE0000} 14.372}}          & 14.409                & \multicolumn{1}{c|}{13.588}                                 & 13.530            & \multicolumn{1}{c|}{17.005}                       & 15.215            & \multicolumn{1}{c|}{{\color[HTML]{FE0000} 62.289}}          & 69.344               \\ \cline{2-10} 
                                 & Quarterly                                     & \multicolumn{1}{c|}{{\color[HTML]{FE0000} 10.67}}           & 10.777                & \multicolumn{1}{c|}{10.909}                                 & 10.875            & \multicolumn{1}{c|}{15.41}                        & 12.696            & \multicolumn{1}{c|}{{\color[HTML]{FE0000} 73.559}}          & 73.585               \\ \cline{2-10} 
                                 & Monthly                                       & \multicolumn{1}{c|}{{\color[HTML]{FE0000} 14.296}}          & 16.650                & \multicolumn{1}{c|}{{\color[HTML]{FE0000} 14.149}}          & 14.431            & \multicolumn{1}{c|}{19.075}                       & 15.210            & \multicolumn{1}{c|}{{\color[HTML]{FE0000} 69.795}}          & 69.80                \\ \cline{2-10} 
                                 & Others                                        & \multicolumn{1}{c|}{{\color[HTML]{FE0000} 5.529}}           & 5.543                 & \multicolumn{1}{c|}{{\color[HTML]{FE0000} 5.853}}           & 5.866             & \multicolumn{1}{c|}{{\color[HTML]{FE0000} 7.424}} & 7.483             & \multicolumn{1}{c|}{98.498}                                 & 98.492               \\ \cline{2-10} 
\multirow{-5}{*}{\textbf{sMAPE}} & \textbf{Average}                              & \multicolumn{1}{c|}{{\color[HTML]{FE0000} \textbf{13.005}}} & \textbf{14.170}       & \multicolumn{1}{c|}{{\color[HTML]{FE0000} \textbf{12.827}}} & \textbf{14.942}   & \multicolumn{1}{c|}{\textbf{17.137}}              & \textbf{14.306}   & \multicolumn{1}{c|}{{\color[HTML]{FE0000} \textbf{71.017}}} & \textbf{72.038}      \\ \hline
                                 & \multicolumn{1}{l|}{\textbf{Difference (\%)}} & \multicolumn{2}{c|}{{\color[HTML]{3166FF} \textbf{8.22}}}                           & \multicolumn{2}{c|}{{\color[HTML]{3166FF} \textbf{14.15}}}                      & \multicolumn{2}{c|}{-}                                                & \multicolumn{2}{c|}{{\color[HTML]{3166FF} \textbf{1.41}}}                          \\ \hline
\textbf{MAPE}                    & Yearly                                        & \multicolumn{1}{c|}{{\color[HTML]{FE0000} 17.928}}          & 19.191                & \multicolumn{1}{c|}{{\color[HTML]{FE0000} 16.709}}          & 16.89             & \multicolumn{1}{c|}{25.554}                       & 19.837            & \multicolumn{1}{c|}{{\color[HTML]{FE0000} 57.533}}          & 61.950               \\ \hline
                                 & Quarterly                                     & \multicolumn{1}{c|}{{\color[HTML]{FE0000} 12.351}}          & 12.871                & \multicolumn{1}{c|}{{\color[HTML]{FE0000} 12.924}}          & 13.052            & \multicolumn{1}{c|}{18.670}                       & 14.969            & \multicolumn{1}{c|}{{\color[HTML]{FE0000} 67.002}}          & 66.971               \\ \hline
                                 & Monthly                                       & \multicolumn{1}{c|}{{\color[HTML]{FE0000} 16.783}}          & 20.144                & \multicolumn{1}{c|}{{\color[HTML]{FE0000} 16.873}}          & 17.635            & \multicolumn{1}{c|}{21.334}                       & 17.972            & \multicolumn{1}{c|}{68.516}                                 & 68.507               \\ \hline
                                 & Others                                        & \multicolumn{1}{c|}{7.76}                                   & 7.750                 & \multicolumn{1}{c|}{{\color[HTML]{FE0000} 10.478}}          & 11.278            & \multicolumn{1}{c|}{13.513}                       & 10.469            & \multicolumn{1}{c|}{{\color[HTML]{FE0000} 64.923}}          & 64.928               \\ \hline
                                 & \textbf{Average}                              & \multicolumn{1}{c|}{{\color[HTML]{FE0000} \textbf{15.532}}} & \textbf{17.560}       & \multicolumn{1}{c|}{{\color[HTML]{FE0000} \textbf{15.568}}} & \textbf{16.046}   & \multicolumn{1}{c|}{\textbf{21.274}}              & \textbf{17.305}   & \multicolumn{1}{c|}{{\color[HTML]{FE0000} \textbf{65.447}}} & \textbf{66.451}      \\ \hline
                                 & \multicolumn{1}{l|}{\textbf{Difference (\%)}} & \multicolumn{2}{c|}{{\color[HTML]{3166FF} \textbf{11.55}}}                          & \multicolumn{2}{c|}{{\color[HTML]{3166FF} \textbf{2.98}}}                       & \multicolumn{2}{c|}{-}                                                & \multicolumn{2}{c|}{{\color[HTML]{3166FF} \textbf{1.51}}}                          \\ \hline
\textbf{MASE}                    & Yearly                                        & \multicolumn{1}{c|}{{\color[HTML]{FE0000} 3.215}}           & 3.218                 & \multicolumn{1}{c|}{{\color[HTML]{FE0000} 3.036}}           & 3.06              & \multicolumn{1}{c|}{{\color[HTML]{FE0000} 3.719}} & 3.798             & \multicolumn{1}{c|}{{\color[HTML]{FE0000} 17.399}}          & 18.11                \\ \hline
                                 & Quarterly                                     & \multicolumn{1}{c|}{{\color[HTML]{FE0000} 1.276}}           & 1.284                 & \multicolumn{1}{c|}{1.315}                                  & 1.302             & \multicolumn{1}{c|}{2.104}                        & 1.561             & \multicolumn{1}{c|}{13.317}                                 & 13.313               \\ \hline
                                 & Monthly                                       & \multicolumn{1}{c|}{{\color[HTML]{FE0000} 1.129}}           & 1.392                 & \multicolumn{1}{c|}{{\color[HTML]{FE0000} 1.122}}           & 1.138             & \multicolumn{1}{c|}{1.75}                         & 1.217             & \multicolumn{1}{c|}{11.169}                                 & 11.168               \\ \hline
                                 & Others                                        & \multicolumn{1}{c|}{{\color[HTML]{FE0000} 3.94}}            & 3.998                 & \multicolumn{1}{c|}{3.782}                                  & 3.754             & \multicolumn{1}{c|}{5.407}                        & 4.937             & \multicolumn{1}{c|}{{\color[HTML]{FE0000} 79.684}}          & 79.686               \\ \hline
                                 & \textbf{Average}                              & \multicolumn{1}{c|}{{\color[HTML]{FE0000} \textbf{1.532}}}  & \textbf{1.916}        & \multicolumn{1}{c|}{{\color[HTML]{FE0000} \textbf{1.742}}}  & \textbf{1.750}    & \multicolumn{1}{c|}{\textbf{2.471}}               & \textbf{1.994}    & \multicolumn{1}{c|}{{\color[HTML]{FE0000} \textbf{16.543}}} & \textbf{16.705}      \\ \hline
                                 & \multicolumn{1}{l|}{\textbf{Difference (\%)}} & \multicolumn{2}{c|}{{\color[HTML]{3166FF} \textbf{20.04}}}                          & \multicolumn{2}{c|}{{\color[HTML]{3166FF} \textbf{0.5}}}                        & \multicolumn{2}{c|}{-}                                                & \multicolumn{2}{c|}{{\color[HTML]{3166FF} \textbf{0.97}}}                          \\ \hline
\textbf{OWA}                     & Yearly                                        & \multicolumn{1}{c|}{{\color[HTML]{FE0000} 0.844}}           & 0.846                 & \multicolumn{1}{c|}{{\color[HTML]{FE0000} 0.798}}           & 0.799             & \multicolumn{1}{c|}{0.988}                        & 0.893             & \multicolumn{1}{c|}{{\color[HTML]{FE0000} 4.309}}           & 4.40                 \\ \hline
                                 & Quarterly                                     & \multicolumn{1}{c|}{{\color[HTML]{FE0000} 0.950}}           & 0.957                 & \multicolumn{1}{c|}{0.975}                                  & 0.969             & \multicolumn{1}{c|}{{\color[HTML]{FE0000} 1.367}} & 1.445             & \multicolumn{1}{c|}{8.196}                                  & 8.195                \\ \hline
                                 & Monthly                                       & \multicolumn{1}{c|}{{\color[HTML]{FE0000} 1.026}}           & 1.232                 & \multicolumn{1}{c|}{{\color[HTML]{FE0000} 1.018}}           & 1.035             & \multicolumn{1}{c|}{1.484}                        & 1.299             & \multicolumn{1}{c|}{7.670}                                  & 7.670                \\ \hline
                                 & Others                                        & \multicolumn{1}{c|}{{\color[HTML]{FE0000} 1.203}}           & 1.214                 & \multicolumn{1}{c|}{1.212}                                  & 1.209             & \multicolumn{1}{c|}{{\color[HTML]{FE0000} 1.634}} & 1.702             & \multicolumn{1}{c|}{22.930}                                 & 22.930               \\ \hline
                                 & \textbf{Average}                              & \multicolumn{1}{c|}{{\color[HTML]{FE0000} \textbf{0.946}}}  & \textbf{1.023}        & \multicolumn{1}{c|}{{\color[HTML]{FE0000} \textbf{0.928}}}  & \textbf{0.935}    & \multicolumn{1}{c|}{\textbf{1.278}}               & \textbf{1.143}    & \multicolumn{1}{c|}{{\color[HTML]{FE0000} \textbf{6.981}}}  & \textbf{7.024}       \\ \hline
                                 & \multicolumn{1}{l|}{\textbf{Difference (\%)}} & \multicolumn{2}{c|}{{\color[HTML]{3166FF} \textbf{7.53}}}                           & \multicolumn{2}{c|}{{\color[HTML]{3166FF} \textbf{0.75}}}                       & \multicolumn{2}{c|}{-}                                                & \multicolumn{2}{c|}{{\color[HTML]{3166FF} \textbf{0.61}}}                          \\ \hline
\end{tabular}
\label{table-resultsM4}
\end{table*}

\begin{figure}[hbt!]
  \begin{subfigure}{0.8\columnwidth}
  \includegraphics[width=\textwidth]{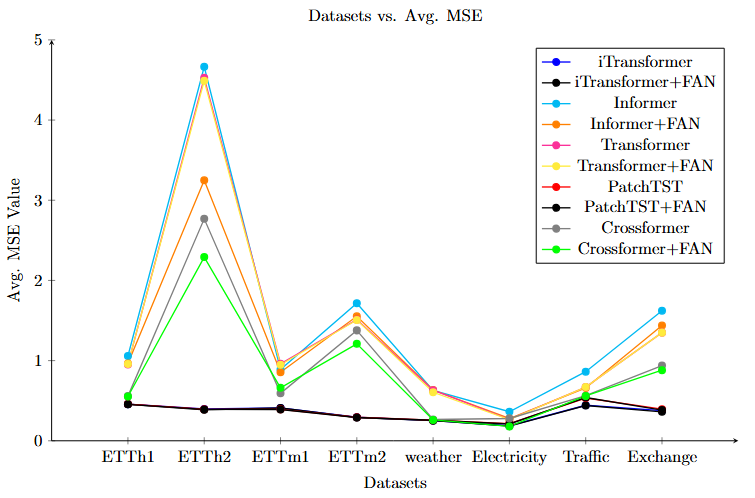}
  \end{subfigure}
  \hfill
  \begin{subfigure}{0.8\columnwidth} 
  \includegraphics[width=\textwidth]{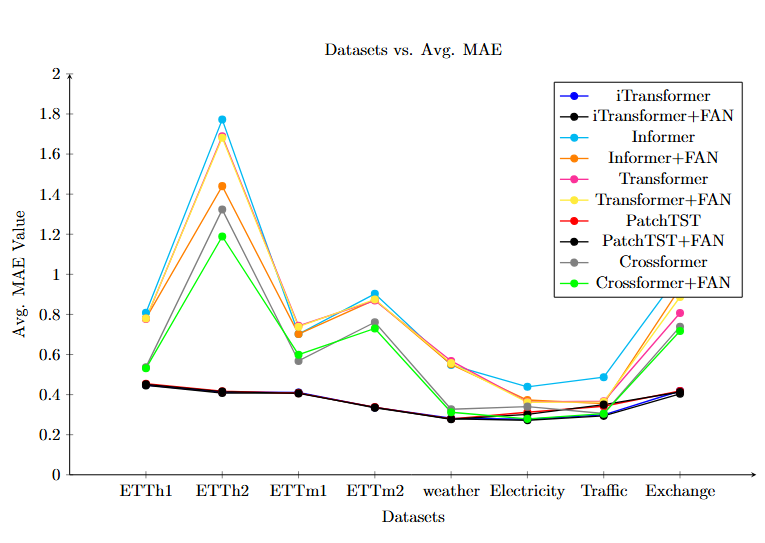} 
  \end{subfigure}
  \caption{Comparison of models efficiency in terms of forecasting}
  \label{fig-comparison}
\end{figure}

\begin{figure}[hbt!]
  \begin{subfigure}{0.49\columnwidth}
  \includegraphics[width=\textwidth]{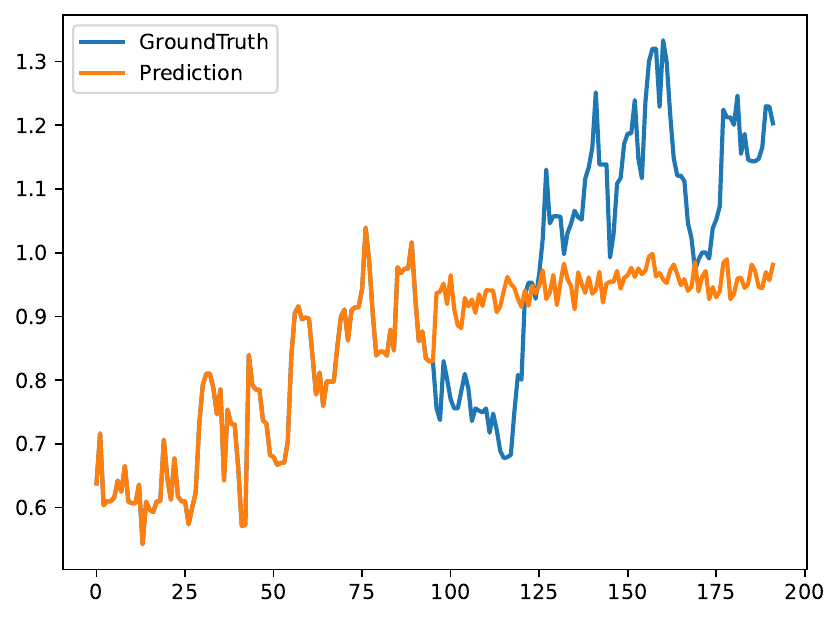}
  \caption{iTransformer} 
  \end{subfigure} 
  \hfill
  \begin{subfigure}{0.49\columnwidth} 
  \includegraphics[width=\textwidth]{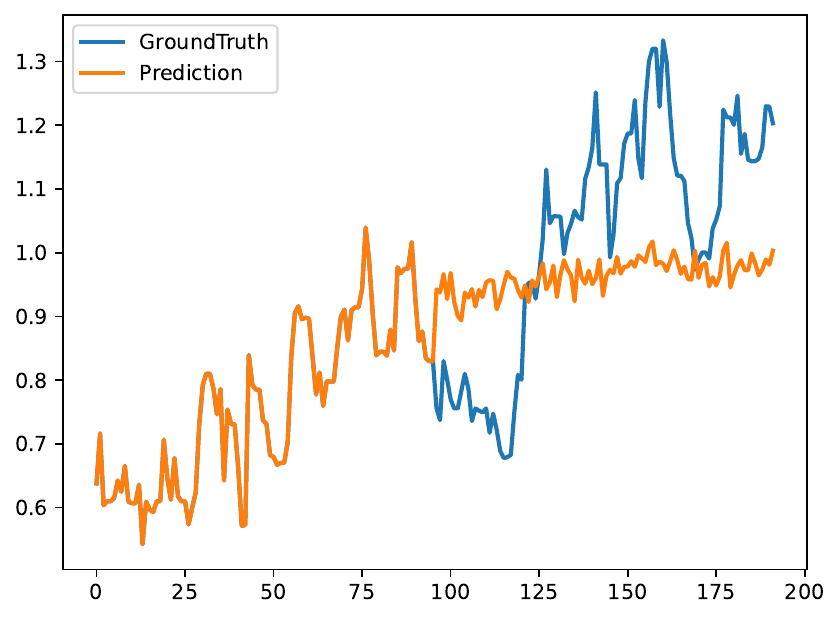} 
  \caption{iTransformer+FAN} 
  \end{subfigure}
  \begin{subfigure}{0.49\columnwidth}
  \includegraphics[width=\textwidth]{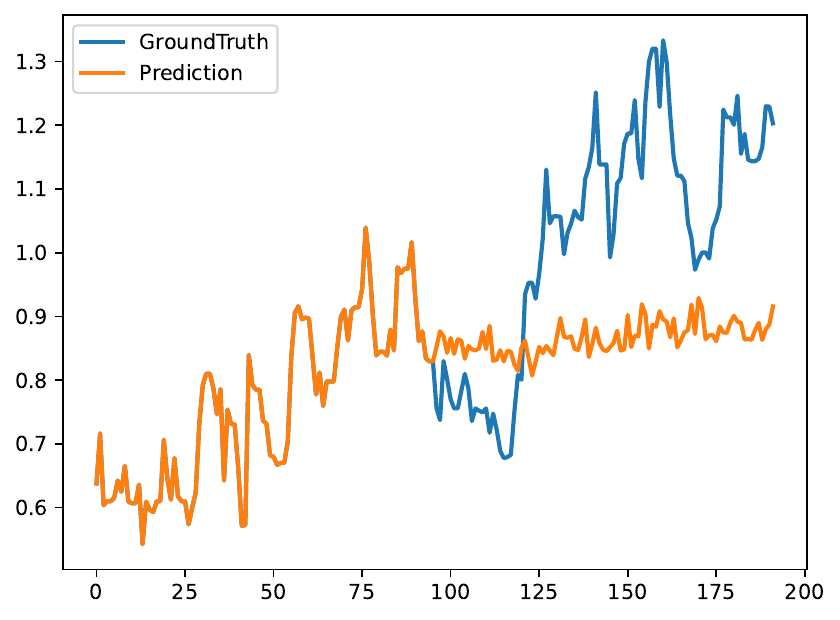}
  \caption{PatchTST}
  \end{subfigure} 
  \hfill 
  \begin{subfigure}{0.49\columnwidth} 
  \includegraphics[width=\textwidth]{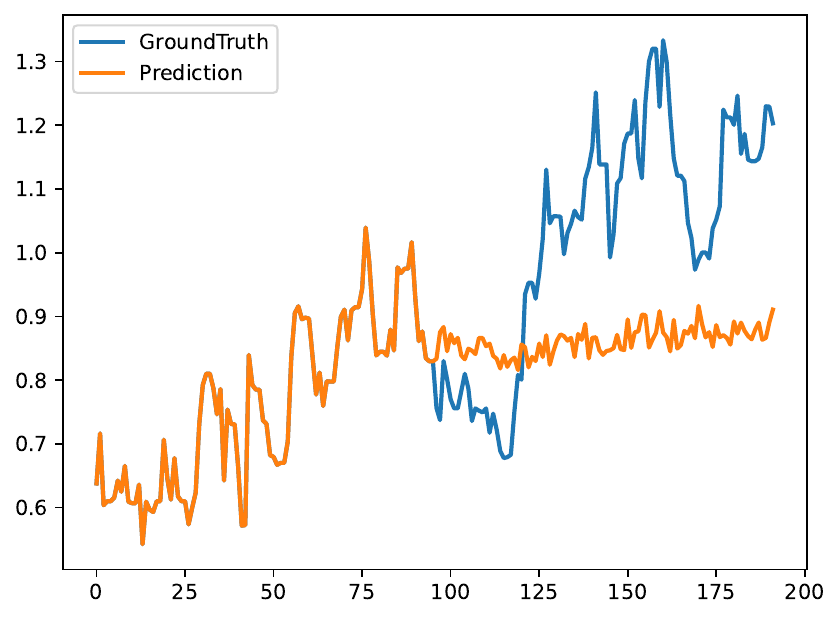} 
  \caption{PatchTST+FAN} 
  \end{subfigure}  
  \hfill
  \begin{subfigure}{0.49\columnwidth} 
  \includegraphics[width=\textwidth]{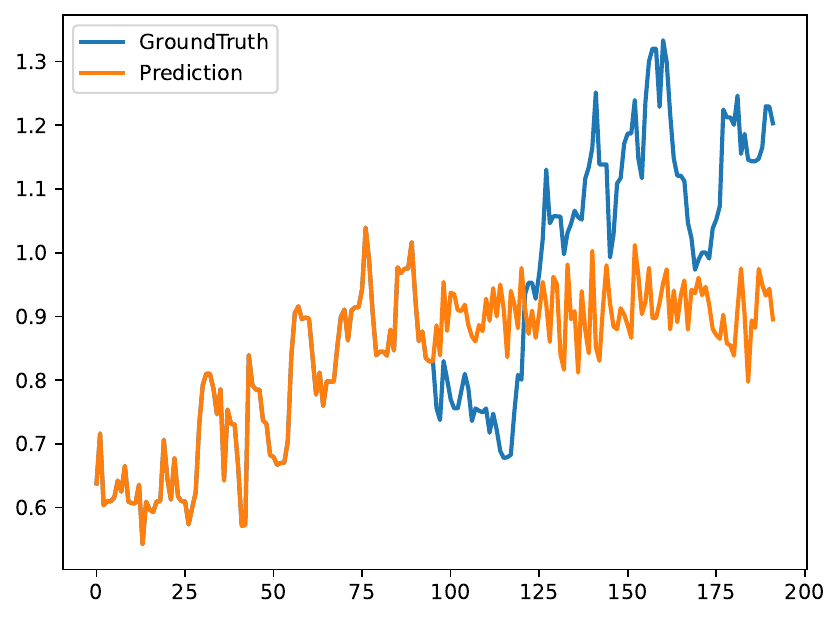} 
  \caption{Crossformer} 
  \end{subfigure}  
  \hfill
  \begin{subfigure}{0.49\columnwidth} 
  \includegraphics[width=\textwidth]{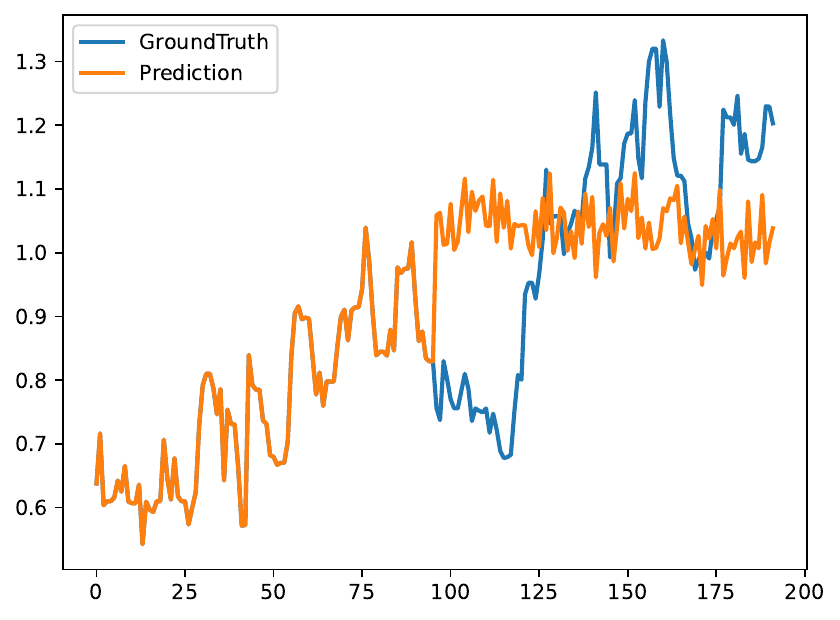} 
  \caption{Crossformer+FAN} 
  \end{subfigure} 
   \hfill
   \begin{subfigure}{0.49\columnwidth} 
  \includegraphics[width=\textwidth]{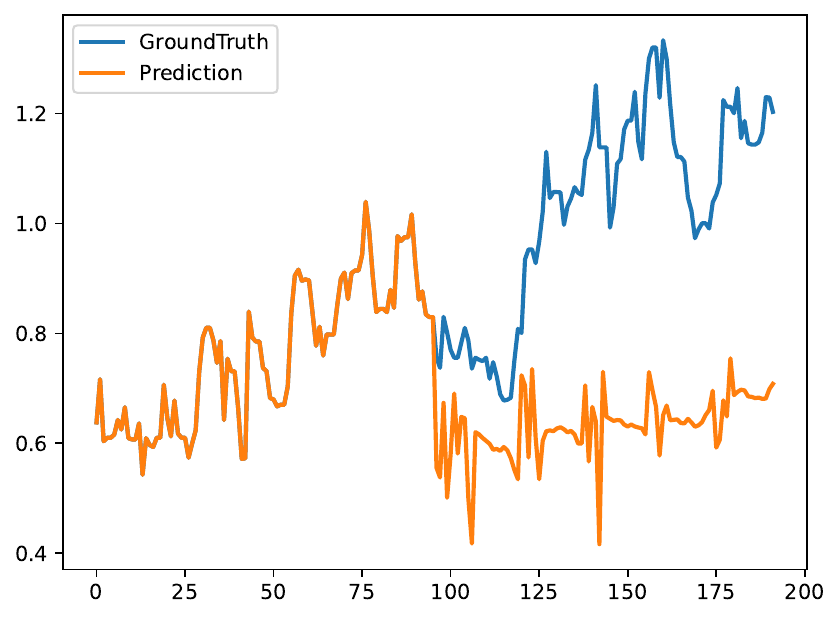} 
  \caption{Informer} 
  \end{subfigure}  
  \hfill
  \begin{subfigure}{0.49\columnwidth} 
  \includegraphics[width=\textwidth]{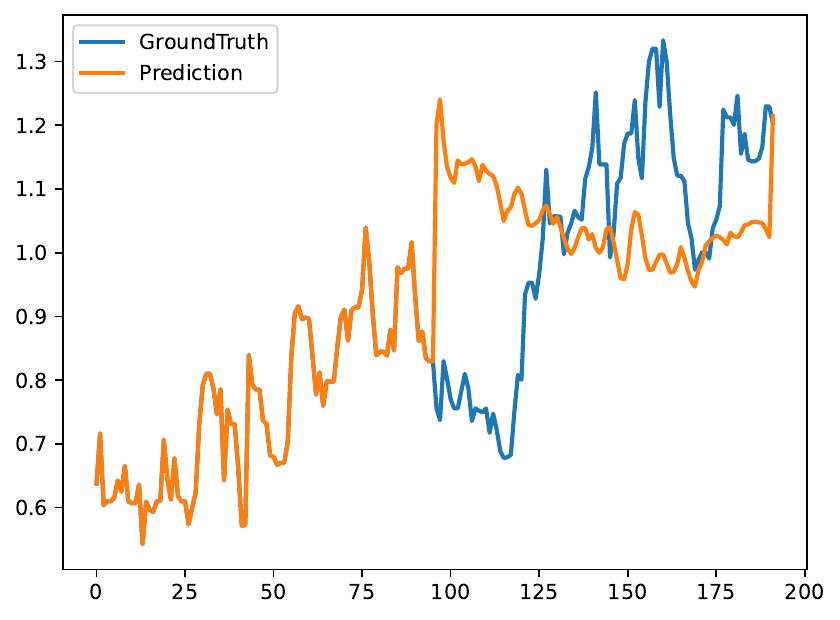} 
  \caption{Informer+FAN} 
  \end{subfigure}  
  \hfill
  \begin{subfigure}{0.49\columnwidth} 
  \includegraphics[width=\textwidth]{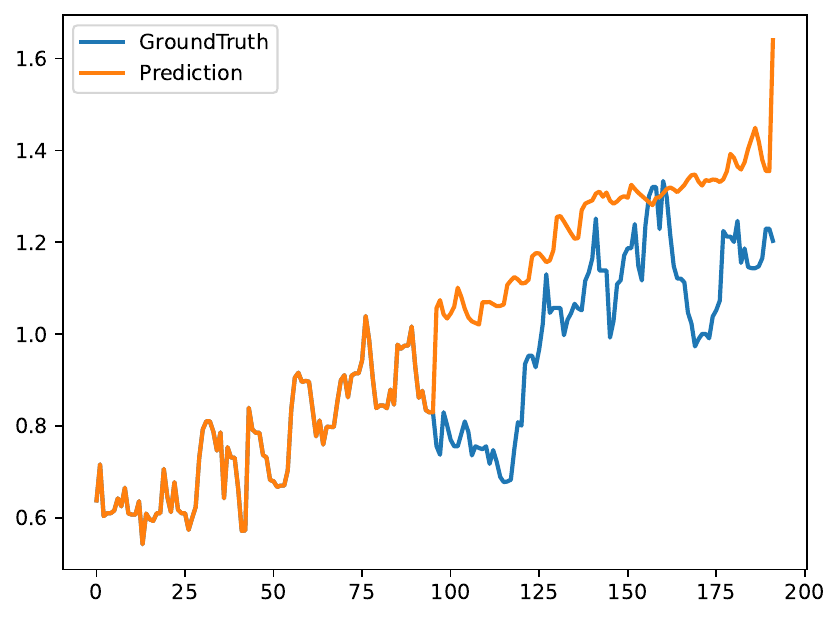} 
  \caption{Transformer} 
  \end{subfigure} 
   \hfill
   \begin{subfigure}{0.49\columnwidth} 
  \includegraphics[width=\textwidth]{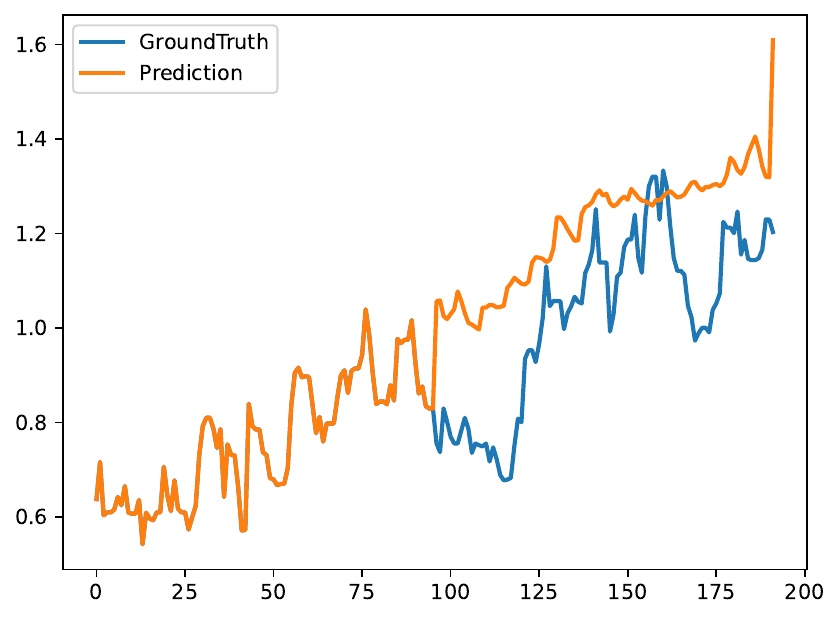} 
  \caption{Transformer+FAN} 
  \end{subfigure}  
  \hfill
   
  \caption{Prediction graphs on Exchange dataset (prediction length:96)}
  \label{predtest1}
\end{figure}

\begin{figure}[hbt!]
  \begin{subfigure}{0.4\columnwidth}
  \includegraphics[width=\textwidth]{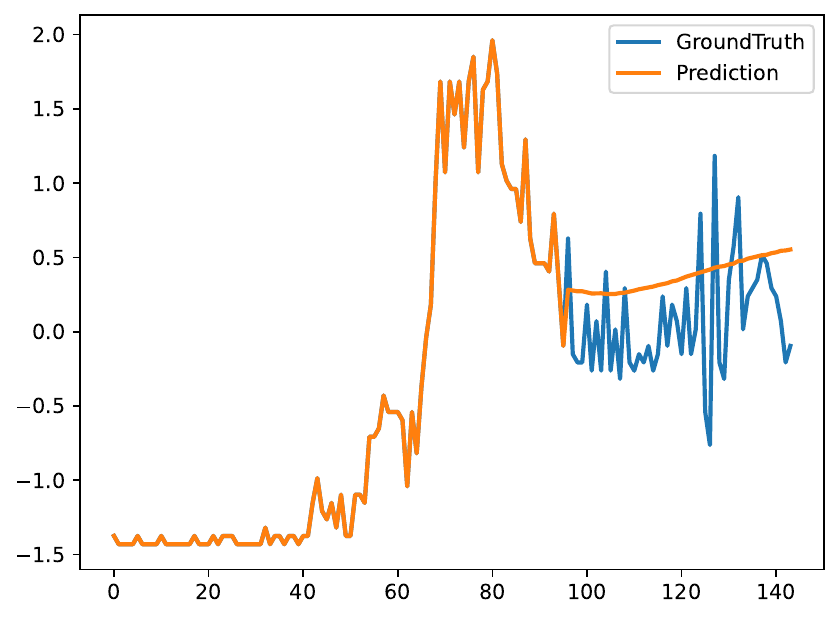}
  \caption{iTransformer+FAN} 
  \end{subfigure} 
  \hfill
  \begin{subfigure}{0.49\columnwidth} 
  \includegraphics[width=\textwidth]{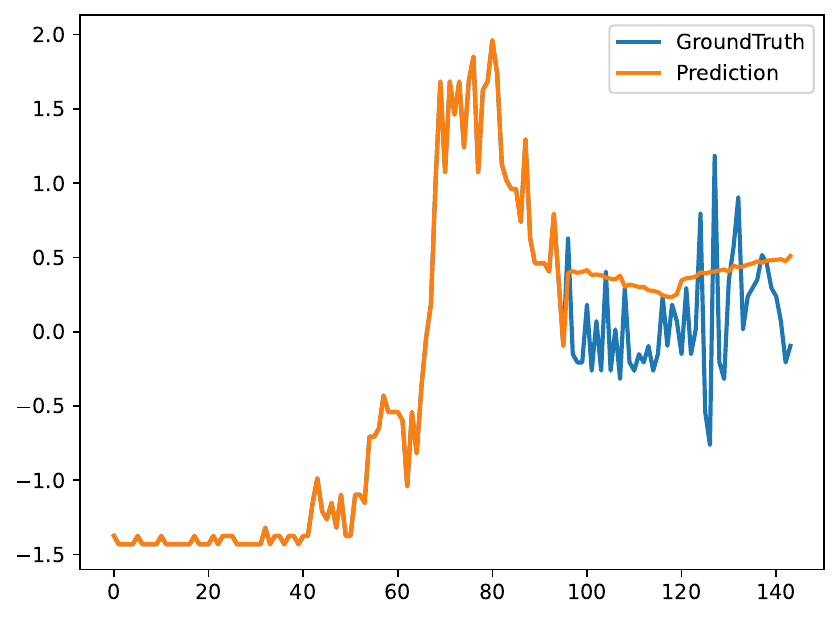} 
  \caption{Crossformer+FAN} 
  \end{subfigure}
  \begin{center}
  \begin{subfigure}{0.49\columnwidth} 
    \includegraphics[width=\textwidth]{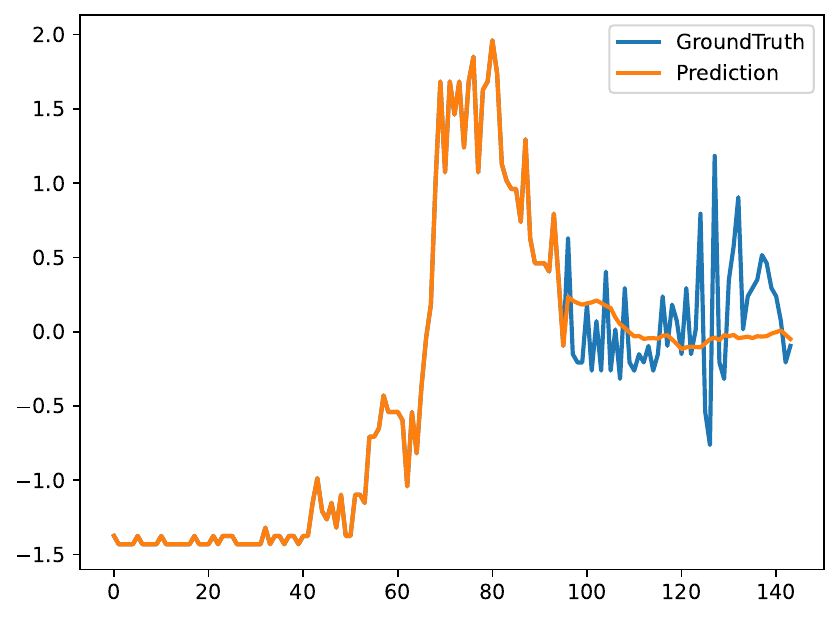} 
    \caption{PatchTST+FAN} 
  \end{subfigure}  
\end{center} 
  \hfill 
  \caption{Prediction graphs on PEMS08 dataset for prediction length 48 for three FANTF approach}
  \label{fig-PEMS08-compare}
\end{figure}

\subsection{Classification}
\begin{figure}[hbt!]
\centering
\includegraphics[scale=0.35]{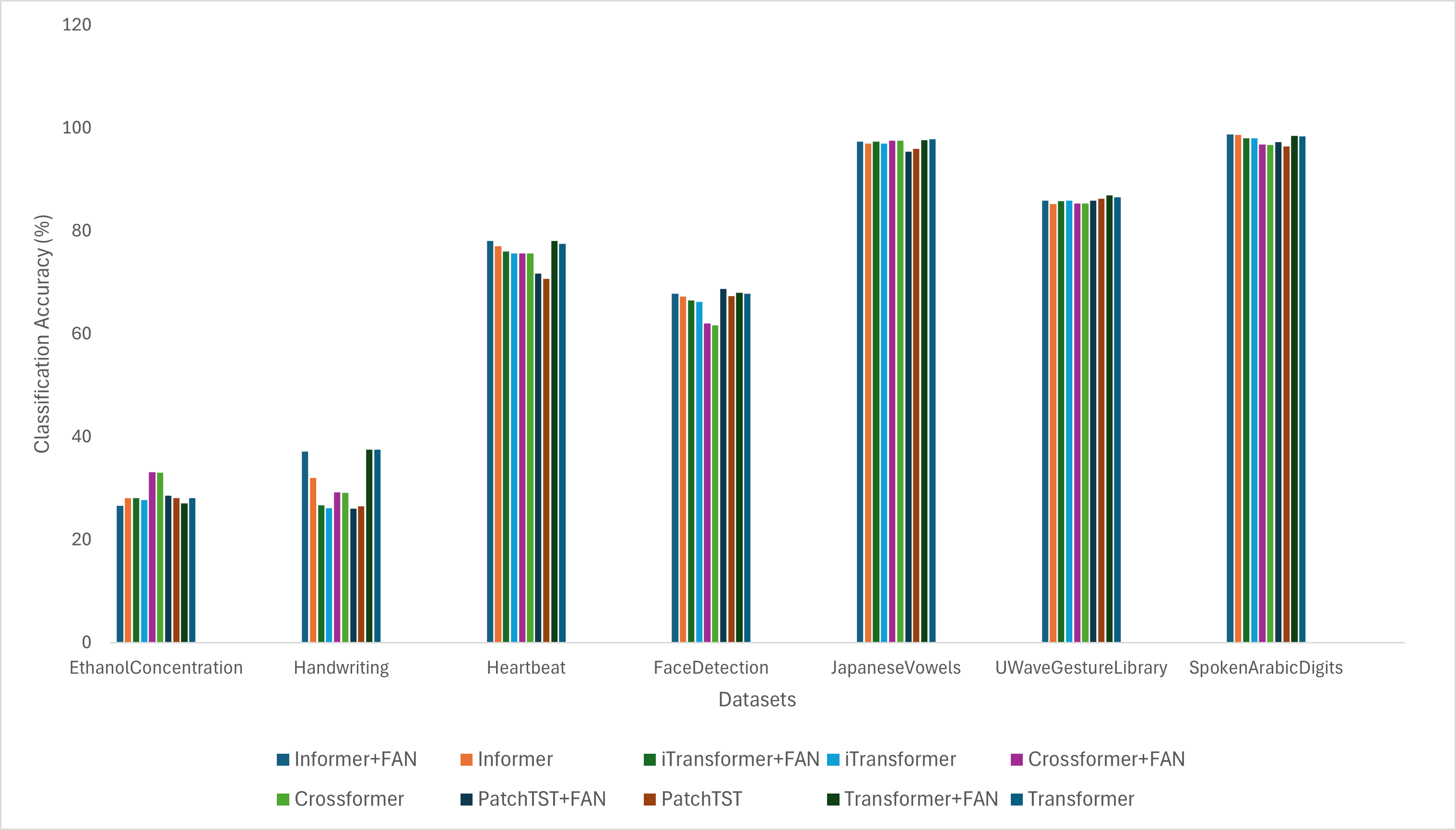}
\caption{Models comparison with accuracy (\%) in classification task on the 7 datasets of UEA.}
\label{FANTF_classification}
\end{figure}

Classifying time series is important for jobs like medical diagnosis and recognition \citep{moody2011physionet}. Sequence-level classification is used in this work to assess the model's capacity to learn high-level representations. Seven multivariate datasets have been specifically chosen from the UEA Time Series Classification Archive \cite{bagnall2018uea}, encompassing diverse applications such as gesture, face, and audio recognition, medical diagnosis through heartbeat tracking, and other real-life applications. Subsequently, we have preprocessed the datasets based on the methodology described in \citep{zerveas2021transformer}, ensuring compatibility with varying sequence lengths across different subsets. As illustrated in Fig. \ref{FANTF_classification}, FANTF demonstrates significant performance improvements in classification accuracy, outperforming conventional state-of-the-art transformers due to FAN's robust noise resiliency, enhanced expressive capability, and adaptive fuzziness mechanism.

\subsection{Anomaly Detection}
Detecting anomalies in monitoring data is critical for effective industrial maintenance. Given that anomalies are often concealed within large-scale datasets, labeling such data becomes a significant challenge. Therefore, our focus is on unsupervised time series anomaly detection, aimed at identifying abnormal time points without labeled data. We evaluate models across five widely used anomaly detection benchmarks: SMD, MSL, SMAP, SWaT, and PSM. The details of these datasets are described in the supplementary section. These datasets span various applications, including service tracking, exploration of space and earth, and systems for water treatment. In line with the pre-processing methods outlined in Anomaly Transformer (2021), we segment the datasets into consecutive, non-overlapping sequences using a sliding window approach. Table \ref{ano} shows that FANTF consistently achieves superior performance in anomaly detection across most cases, surpassing state-of-the-art Transformer-based models such as Informer, iTransformer, PatchTST, and others.

\begin{table*}[hbt!]
\centering
\scriptsize
\tiny
\captionsetup{justification=centering}
\caption{Anomaly detection task. We calculate the F1-score (as \%) for each dataset. A higher value of F1-score indicates better performance. The difference between the results is shown in \% where the blue colours represent improvements of FANTF over existing models.}
\begin{tabular}{|c|cc|cc|cc|cc|cc|}
\hline
Models                                         & \multicolumn{1}{c|}{Informer+FAN}                & Informer       & \multicolumn{1}{c|}{iTransformer+FAN}   & iTransformer    & \multicolumn{1}{c|}{Crossformer+FAN}    & Crossformer     & \multicolumn{1}{c|}{PatchTST+FAN}      & PatchTST         & \multicolumn{1}{c|}{Transformer+FAN}   & Transformer      \\ \hline
SMD                                            & \multicolumn{1}{c|}{{\color[HTML]{000000} 71.2}} & 71.1           & \multicolumn{1}{c|}{82.3}               & 82.1            & \multicolumn{1}{c|}{73.6}               & 73.6            & \multicolumn{1}{c|}{84.3}              & 84.1             & \multicolumn{1}{c|}{71.2}              & 71.1             \\ \hline
\multicolumn{1}{|l|}{\textbf{Difference (\%)}} & \multicolumn{2}{c|}{{\color[HTML]{3166FF} \textbf{0.14}}}         & \multicolumn{2}{c|}{{\color[HTML]{3166FF} \textbf{0.24}}} & \multicolumn{2}{c|}{{\color[HTML]{3166FF} \textbf{0}}}    & \multicolumn{2}{c|}{{\color[HTML]{3166FF} \textbf{0.24}}} & \multicolumn{2}{c|}{{\color[HTML]{3166FF} \textbf{0.14}}} \\ \hline
MSL                                            & \multicolumn{1}{c|}{81.8}                        & 81.2           & \multicolumn{1}{c|}{72.8}               & 72.4            & \multicolumn{1}{c|}{81.3}               & 80.8            & \multicolumn{1}{c|}{78.5}              & 79.1             & \multicolumn{1}{c|}{81.6}              & 80.8             \\ \hline
\multicolumn{1}{|l|}{\textbf{Difference (\%)}} & \multicolumn{2}{c|}{{\color[HTML]{3166FF} \textbf{0.74}}}         & \multicolumn{2}{c|}{{\color[HTML]{3166FF} \textbf{0.28}}} & \multicolumn{2}{c|}{{\color[HTML]{3166FF} \textbf{0.62}}} & \multicolumn{2}{c|}{{\color[HTML]{3166FF} \textbf{-}}}    & \multicolumn{2}{c|}{{\color[HTML]{3166FF} \textbf{0.99}}} \\ \hline
SMAP                                           & \multicolumn{1}{c|}{73.3}                        & 68.9           & \multicolumn{1}{c|}{67.1}               & 66.9            & \multicolumn{1}{c|}{67.2}               & 67.5            & \multicolumn{1}{c|}{67.2}              & 67.2             & \multicolumn{1}{c|}{72.8}              & 73.0             \\ \hline
\multicolumn{1}{|l|}{\textbf{Difference (\%)}} & \multicolumn{2}{c|}{{\color[HTML]{3166FF} \textbf{6.38}}}         & \multicolumn{2}{c|}{{\color[HTML]{3166FF} \textbf{0.24}}} & \multicolumn{2}{c|}{{\color[HTML]{3166FF} \textbf{-}}}    & \multicolumn{2}{c|}{{\color[HTML]{3166FF} \textbf{0}}}    & \multicolumn{2}{c|}{{\color[HTML]{3166FF} \textbf{-}}}    \\ \hline
SWaT                                           & \multicolumn{1}{c|}{81.5}                        & 80.9           & \multicolumn{1}{c|}{92.8}               & 92.7            & \multicolumn{1}{c|}{90.2}               & 90.2            & \multicolumn{1}{c|}{92.4}              & 92.2             & \multicolumn{1}{c|}{81.6}              & 81.5             \\ \hline
\multicolumn{1}{|l|}{\textbf{Difference (\%)}} & \multicolumn{2}{c|}{{\color[HTML]{3166FF} \textbf{0.74}}}         & \multicolumn{2}{c|}{{\color[HTML]{3166FF} \textbf{0.14}}} & \multicolumn{2}{c|}{{\color[HTML]{3166FF} \textbf{0}}}    & \multicolumn{2}{c|}{{\color[HTML]{3166FF} \textbf{0.28}}} & \multicolumn{2}{c|}{{\color[HTML]{3166FF} \textbf{0.14}}} \\ \hline
PSM                                            & \multicolumn{1}{c|}{90.7}                        & 90.7           & \multicolumn{1}{c|}{95.5}               & 95.5            & \multicolumn{1}{c|}{92.7}               & 92.7            & \multicolumn{1}{c|}{96.9}              & 96.8             & \multicolumn{1}{c|}{90.9}              & 90.9             \\ \hline
\multicolumn{1}{|l|}{\textbf{Difference (\%)}} & \multicolumn{2}{c|}{{\color[HTML]{3166FF} \textbf{0}}}            & \multicolumn{2}{c|}{{\color[HTML]{3166FF} \textbf{0}}}    & \multicolumn{2}{c|}{{\color[HTML]{3166FF} \textbf{0}}}    & \multicolumn{2}{c|}{{\color[HTML]{3166FF} \textbf{0.14}}} & \multicolumn{2}{c|}{{\color[HTML]{3166FF} \textbf{0}}}    \\ \hline
\textbf{Avg F1(\%)}                            & \multicolumn{1}{c|}{\textbf{79.7}}               & \textbf{78.56} & \multicolumn{1}{c|}{\textbf{82.1}}      & \textbf{81.92}  & \multicolumn{1}{c|}{\textbf{81.0}}      & \textbf{80.96}  & \multicolumn{1}{c|}{\textbf{83.86}}    & \textbf{83.92}   & \multicolumn{1}{c|}{\textbf{79.62}}    & \textbf{79.46}   \\ \hline
\end{tabular}
\label{ano}
\end{table*}

\section{Conclusions}
\label{conclusions}
In this paper, a fuzzy attention network-based transformer (FANTF), an approach for multivariate time series analysis has been discussed. FANTF is an extended approach which is designed a new attention network with the help of a learnable fuzziness technique. Finally, it embedded itself as an attention module under the full attention layer of the existing time series transformers. It consistently delivers significant performance enhancements across a variety of benchmark transformers and datasets, highlighting its versatility and robustness for long-term and short-term forecasting, classification, and anomaly detection tasks. Additionally, FANTF has shown state-of-the-art runtime efficiency compared to other models. Future work will involve testing this extended approach with large language models (LLM) for time series analysis.

\section*{Acknowledgements}
This work was partially supported by the \textit{Resurssmarta Processor (RSP)}, the Wallenberg AI, Autonomous Systems and Software Program (WASP), and the Wallenberg Initiative Materials Science for Sustainability (WISE), all funded by the Knut and Alice Wallenberg Foundation.
 
\clearpage

\bibliographystyle{unsrt}
\bibliography{reference}

\vfill

\end{document}